\documentclass[runningheads]{llncs}
\usepackage[T1]{fontenc}
\usepackage{graphicx}

\usepackage{cite}
\usepackage{float}
\usepackage{multirow}
\usepackage{rotating}
\usepackage{adjustbox}

\usepackage{xurl}

\usepackage[caption=false,font=footnotesize]{subfig}

\usepackage{xcolor}%

\usepackage{soul} 
\begin{document}
%
\title{An approach with Visual and Tabular Mamba to multimodal medical data using Mixed Fusion}
%
%
%
\titlerunning{An approach with Mamba to multimodal medical data using Mixed Fusion}
%

\author{Anonymous Author\inst{1}}
\authorrunning{A. Author}

\author{Matheus B. Rocha\inst{1,2} \and
Gustavo B. Dettogni \inst{1} \and
Renato A. Krohling\inst{1,2}}
\authorrunning{M. Rocha, G. Dettogni and R. Krohling}

%


\institute{Labcin - Nature Inspired Computing Lab, Federal University of Esp\'irito Santo, Vit\'oria, Brazil \and
PPGI - Graduate Program in Computer Science, Federal University of Esp\'irito Santo, Vit\'oria, Brazil\\
\email{matheusbecali@gmail.com}, \email{gustavo.dettogni@edu.ufes.br}, \email{krohling.renato@gmail.com}
}

\maketitle              
\begin{abstract}
This article presents a complementary approach for integrating multimodal medical data in cancer classification, based on state space models represented by the Mamba architecture. To this end, a mixed multimodal fusion architecture, called Mixed Fusion, was employed and developed to enhance the interpretability of the decision-making process. The proposed approach explores two variants of Mamba: one dedicated to visual processing, responsible for classifying the lesion image and generating probabilities associated with the target classes, and another focused on tabular processing, which uses these probabilities together with clinical and/or sociodemographic data to produce the final diagnosis. The experiments were conducted on two medical datasets: PAD-UFES-20, composed of clinical images and information associated with skin lesions, and NDB-UFES, consisting of histopathological images and sociodemographic data related to oral cancer. The results indicate slightly lower performance in balanced accuracy, compared with Transformer-based approaches, on PAD-UFES-20, and superior performance on NDB-UFES. Additionally, substantial gains were observed in the recall metric. Furthermore, the adoption of the Mixed Fusion architecture enables the application of the Shapley Additive Explanations (SHAP) method, increasing the interpretability of the results. These findings indicate that Mamba-based models constitute a suitable alternative for multimodal classification in medical data, especially in scenarios in which sensitivity is a relevant requirement.

\keywords{Mamba, Multimodal fusion, Artificial intelligent, Interpretable Deep Learning, Cancer diagnosis}
\end{abstract}
\section{Introduction}
\label{sec:intro}

Cancer has become established as one of the leading causes of mortality in the 21st century, as indicated by recent estimates from the International Agency for Research on Cancer (IARC), a specialized agency of the World Health Organization (WHO) \cite{who2024cancer}. Among the most incident cancers, skin cancer stands out due to its high frequency and the need for effective strategies for early detection. In Brazil, recent estimates from the National Cancer Institute (INCA) indicate approximately 781 thousand new cases per year for the period from 2026 to 2028 \cite{inca2026estimativa}. In the context of oral cancer, early detection is crucial for prognosis, since lesions identified at early stages present substantially higher survival rates than those diagnosed late, and anamnesis together with a thorough clinical examination performed by the dentist constitute essential steps for the initial identification of the disease and, consequently, for improving clinical outcomes \cite{grafton2019diagnosis}.

Computer-aided diagnosis (CAD) systems have been developed with the aim of optimizing, standardizing, and accelerating the analysis of medical images, providing support for more accurate cancer identification \cite{ilhan2020improving}. Over the past decades, several CAD systems have been developed to address the challenges of recognizing different types of cancer, such as skin cancer~\cite{takiddin2021artificial,das2021artificial} and oral cancer~\cite{abdul2022classification, amin2021histopathological}. These systems stand out for integrating clinical information and features extracted from images~\cite{Zhou2021FusionMAT, pacheco2020impact, delima2023importance}. This approach, known as multimodal data fusion, has emerged as a promising strategy in the healthcare field, enabling more accurate diagnoses, reducing interobserver variability, and increasing process standardization~\cite{Kline2022multimodalhealth}. The integration of multiple data sources can significantly improve the performance of these systems, making it an essential tool for advancing the diagnosis of cancers such as skin and oral cancer~\cite{pacheco2020impact, zheng2022application, LimaHyperbolic2025}.


In this context, the effectiveness of multimodal systems is directly related to their ability to extract informative visual representations. The use of neural networks has become the standard for visual feature extraction in several tasks, including medical image analysis. Architectures based on deep convolutional neural networks (CNNs) have been and continue to be widely used for this purpose \cite{Jia2024,delima2022exploring}. With the aim of capturing long-range dependencies and broader global context, transformer-based models, such as Vision Transformers (ViTs), have stood out by achieving comparable or superior performance in several skin lesion classification tasks \cite{vitmelanoma2025,delima2022exploring}. More recently, Mamba-based architectures have begun to emerge in medical applications, especially for skin cancer classification \cite{dermamamba2025,yue2024medmamba}.

However, although these models exhibit high predictive power, data integration is predominantly performed at the intermediate stage of the classification process, an approach known as middle fusion. This strategy tends to limit model transparency, posing a critical challenge in the context of medical diagnosis, in which trust, accountability, and understanding of errors are fundamental requirements \cite{Rezk2023, Song2023}. In the case of deep neural networks, the increasing complexity of the architectures, often characterized by a large number of parameters, imposes significant challenges to interpretability, since the pursuit of higher accuracy has resulted in highly complex, opaque, and difficult-to-understand models \cite{yao2024survey}. This trade-off between predictive performance and interpretability has been widely studied, and the ideal balance remains an open field of research \cite{xua2024interpretability,Assis2024}. In light of this limitation, Lima et al. \cite{Lima2026} proposed an approach known as Mixed Fusion, a hybrid approach for multimodal data fusion that enables the interpretability of the predictions produced by the model.

This study proposes the integration of Mamba architectures applied to visual and tabular data, in combination with the Mixed Fusion approach, with the aim of achieving predictive performance close to the state of the art while also providing explainability for the predictions. This strategy seeks to enable medical professionals to understand the factors influencing the model's decisions, thereby fostering greater confidence in its clinical application.

The main contributions of this work are as following:
\begin{itemize}
    \item We propose a complementary approach based on a Mamba-inspired pipeline, combined with a mixed fusion architecture to integrate visual and clinical information.
    \item We conduct experiments on the PAD-UFES-20 and NDB-UFES datasets using different Mamba architecture variants for visual data (MambaVision and VMamba) and tabular data (Mambular and MambAttention), achieving performance close to the baseline and relevant gains in Recall.
\end{itemize}

The structure of the article is presented in the following: Section 2 describes the functioning of Mamba and provides an overview of the multimodal data fusion architecture, which integrates images with clinical and demographic data. Section 3 presents and discusses the experimental results obtained. Finally, Section 4 presents the final conclusions of the study and indicates possible directions for future work.



\section{Background}
\label{sec:background}


As a result of the growing popularity of Large Language Models (LLMs) and, more broadly, foundation models in both academia and among the general public, several innovative proposals have emerged with the aim of reducing one of the main bottlenecks of these technologies, namely the transformer and its highly computationally expensive attention blocks. Although some of these solutions have shown promising results, especially with regard to reducing computational and memory costs \cite{TransformersSurvey2020}, many of them end up involving trade-offs between performance and efficiency, resulting in losses in expressiveness or difficulties in generalization. In this context, one implementation has received particular attention for its balance between performance and representational capacity, namely the Mamba architecture \cite{gu2024mamba}.

Inspired by state space models (SSMs), originally proposed in the 1960s \cite{OriginalSSM1960}, as well as by recurrent (RNN) and convolutional (CNN) architectures, Mamba adopts its own variant of these models, called the Selective State Space Model. This formulation makes it possible to achieve performance comparable to that of Transformers while maintaining linear scalability with respect to the input sequence length. Due to the growing relevance of the Mamba architecture, two derived models specifically designed for image processing have been proposed in the literature in parallel: Vision Mamba (ViM) \cite{VisionMamba2024} and VMamba \cite{VMamba2024}.

The technology proposed by ViM seeks to address two limitations present in the original Mamba model when dealing with visual data: unidirectional modeling and lack of positional awareness. This improvement is achieved by replacing the original Mamba block, designed for processing one-dimensional data, with the Vision Mamba block, which is adapted for two-dimensional data. The new block incorporates a bidirectional SSM capable of providing a global and contextual representation guided by the data, together with positional embeddings that make the interpretation spatially aware.

VMamba, in turn, replaces the original one-dimensional core with a block called Visual State Space (VSS), which employs the 2D-Selective-Scan (SS2D) mechanism. This module incorporates a two-dimensional kernel capable of capturing global context in images and is composed of three main stages: cross-scan, selective scanning using Mamba blocks, and cross-merge. Despite achieving expressive accuracy results on ImageNet with its more compact model, VMamba showed low throughput and high memory overhead.

Notably, the bidirectional SSM employed in Vision Mamba introduces additional computational overhead, impacting both training and inference time, while also limiting the preservation of global context when combining information from both directions. In this context, MambaVision \cite{MambaNvidia2025} emerges as a hybrid architecture that combines attention mechanisms with a one-dimensional state space model through a reformulation of the Mamba block. This architecture was designed to outperform Vision Mamba on benchmarks such as ImageNet-1K while preserving its effectiveness in visual processing.

Another Mamba-inspired architecture used in this study was Mamba Tabular, or Mambular \cite{mambular2025}, which was specifically developed for tabular data processing. In this context, it is important to note how neural network-based models may face difficulties when handling this type of data, especially in scenarios characterized by the presence of missing values and the heterogeneity of attribute types. In addition, the recurring need for extensive preprocessing steps constitutes an additional factor that may hinder their application.

The Mambular architecture is described as follows. Initially, the attributes are partitioned into categorical and numerical variables. The categorical variables are subjected to encoding and vector embedding steps, while the numerical variables are transformed using Periodic Linear Encoding (PLE) \cite{PLE2022}, with the aim of capturing more expressive patterns than those obtained through a simple linear transformation. Next, the transformed attributes are processed by a one-dimensional convolutional layer and an SSM. The final representation is obtained by combining the stacked hidden states through weighted multiplication and summation, followed by the addition of a scaled adjustment term derived from the input. Subsequently, this representation is passed through a linear layer and finally forwarded to the output layer, whose configuration depends on the task under consideration. In this work, a classification layer was used due to the nature of the problem investigated. Similarly to MambaVision, a hybrid architecture for tabular data was also developed, called MambAttention \cite{mambattention2024}, which combines attention blocks in parallel with state space models.


\section{Methodology}
\label{sec:methgy}

    
    Proposed by Lima et al. \cite{Lima2026}, Mixed Fusion consists of a hybrid fusion architecture inspired by the diagnostic decision-making process adopted by specialists. In contrast to traditional approaches, such as early fusion, feature-level fusion, and late fusion, Mixed Fusion combines feature-level information with probabilities generated in an earlier stage of the classification process. This conception is based on the observation that specialists often begin the analysis with a preliminary decision based exclusively on visual information and then incorporate complementary data obtained through anamnesis to refine the diagnosis before reaching the final result. In this architecture, the image backbone produces class probabilities based solely on the input image. Subsequently, these probabilities, generated by the internal classifier, are concatenated with complementary data and processed by an external classifier. 

\begin{figure}[ht]
        \centerline{\includegraphics[scale=0.3]{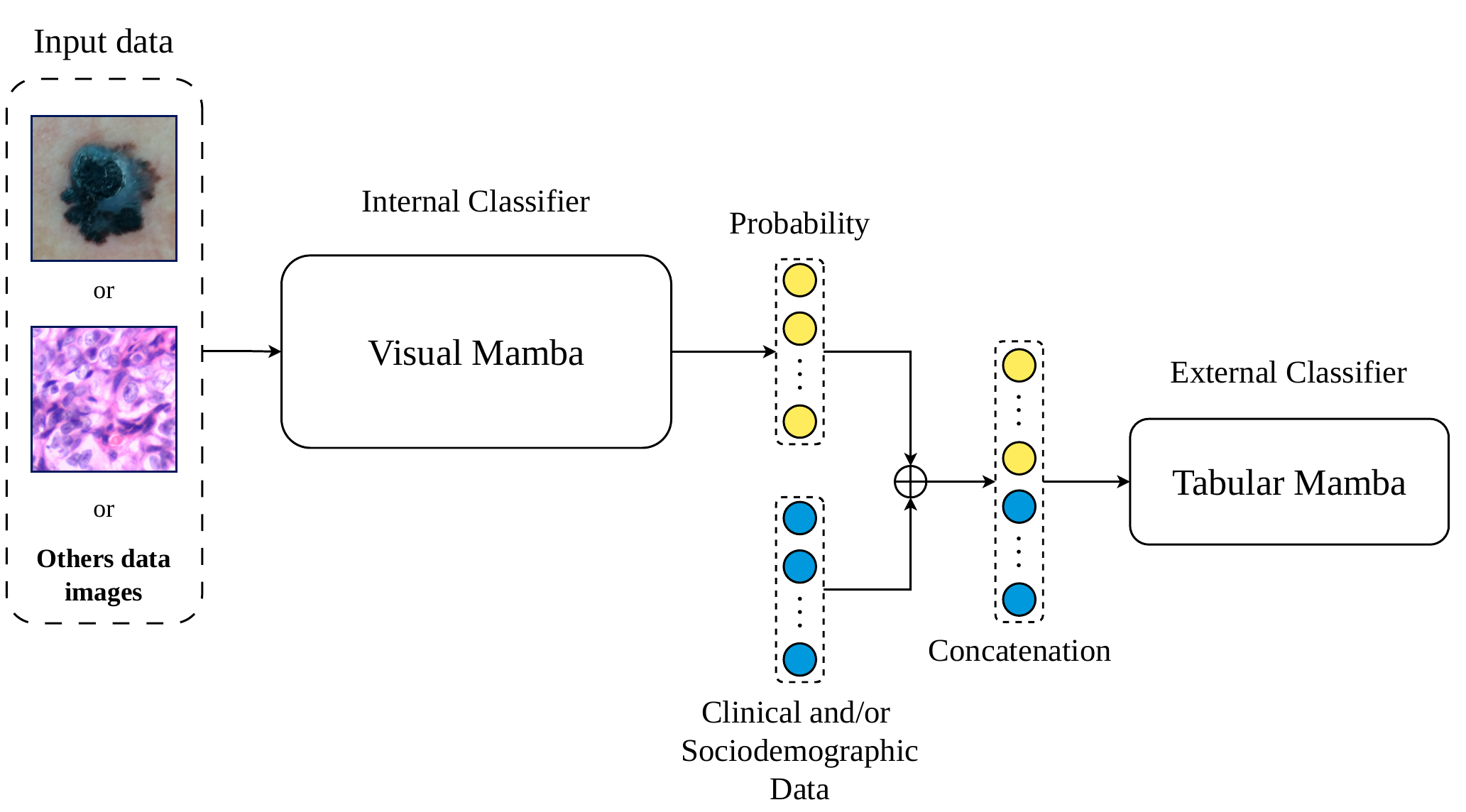}}
        \caption{Mixed fusion architecture. The images are initially processed by a feature extractor and then passed through a classifier that generates class probabilities. These probabilities are combined with clinical data, resulting in a final classification step}
        \label{fig:mixedfusion_arch}
    \end{figure}

   In the original formulation of Mixed Fusion, the internal and external classifiers are based on convolutional architectures and classical machine learning methods, respectively. In this work, both classifiers were replaced by Mamba-based models, given the linear complexity and ability to model long-range dependencies of this family of architectures. For the internal classifier, two Mamba models designed for visual data processing were adopted: MambaVision \cite{MambaNvidia2025} and VMamba \cite{VMamba2024}. For the external classifier, two Mamba-based models aimed at tabular data processing were adopted: Mambular \cite{mambular2025} and MambAttention \cite{mambattention2024}, as illustrated in Figure~\ref{fig:mixedfusion_arch}.


    In contrast to conventional approaches, the Mixed Fusion architecture makes it possible to integrate the SHapley Additive exPlanations (SHAP) method \cite{scott2017unified}, allowing the analysis of the impact of each individual attribute on the model predictions and the interpretation of its global behavior. SHAP is a post hoc explanation method based on an additive model, proposed as a unified measure of feature importance and grounded in cooperative game theory, in which each attribute is considered an agent that contributes to the model prediction. SHAP values derive from the adaptation of Shapley values to the conditional expectation function of a machine learning model, such that the Shapley value quantifies the contribution of each input attribute to the difference between the model prediction and the average value of the predictions.



\section{Experiments}
\label{sec:results}

\subsection{PAD-UFES-20 Dataset}

    The experiments were conducted using datasets related to skin and oral lesions. The first dataset analyzed in this study is PAD-UFES-20~\cite{pacheco2020pad}, which combines clinical images and clinical data from patients who are mostly descendants of European immigrants and predominantly classified as having fair skin phototypes. A substantial portion of these patients are or were rural workers, with prolonged sun exposure and, in many cases, without adequate protection.

\begin{figure}[!htbp]
        \centering
        \subfloat[ACK.]{
            \includegraphics[height=3.25cm, width=3.25cm]{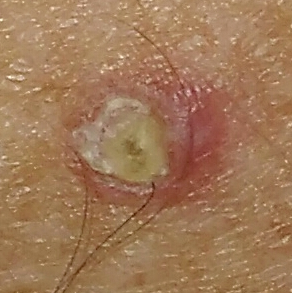}
        }
        \hspace*{0.05cm}
        \subfloat[SEK.]{
            \includegraphics[height=3.25cm, width=3.25cm]{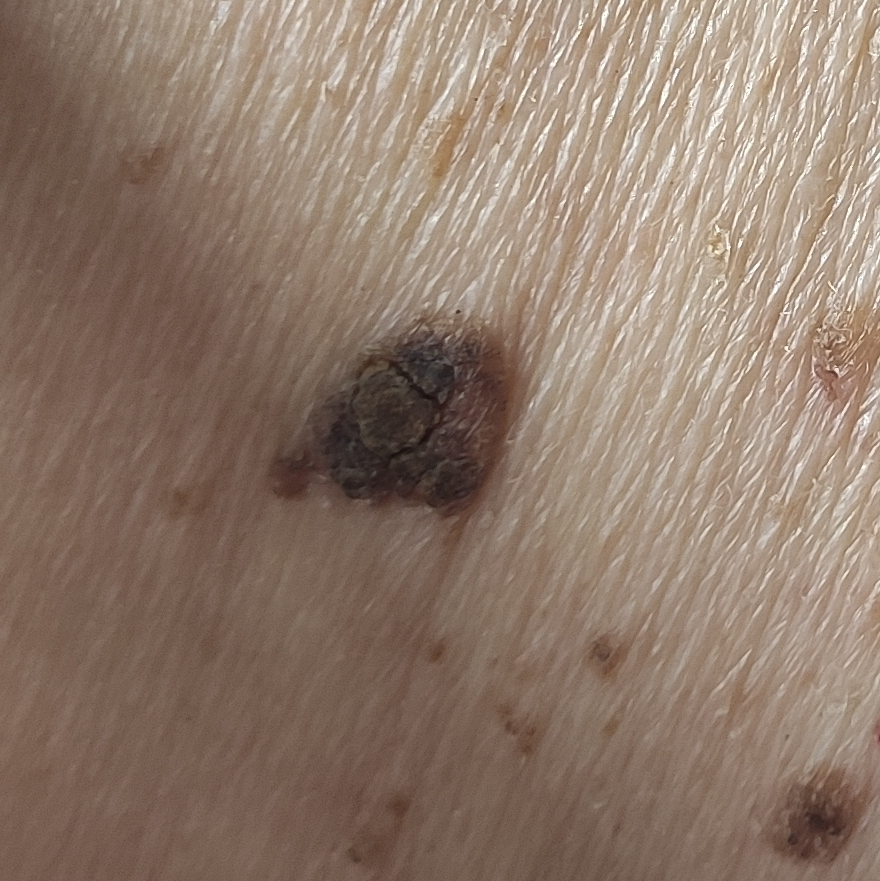}
        }
        \hspace*{0.05cm}
        \subfloat[NEV.]{
            \includegraphics[height=3.25cm, width=3.25cm]{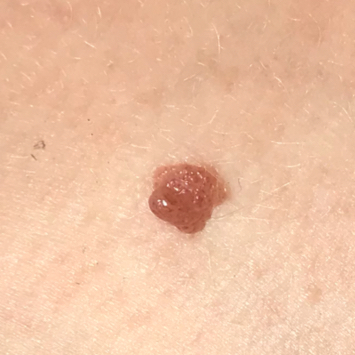}
        }
        \hfill
        \subfloat[BCC.]{
            \includegraphics[height=3.25cm, width=3.25cm]{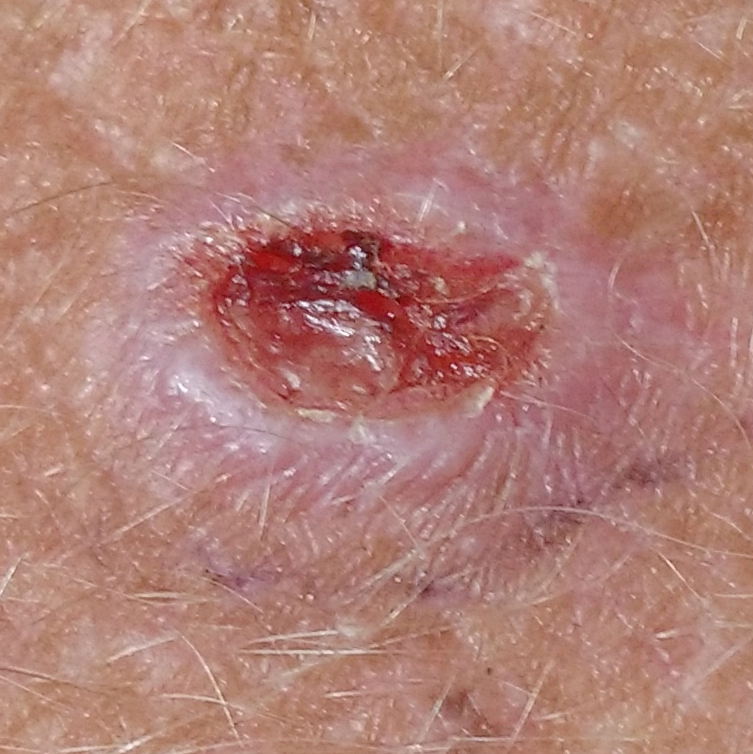}
        }
        \hspace*{0.05cm}
        \subfloat[SCC.]{
            \includegraphics[height=3.25cm, width=3.25cm]{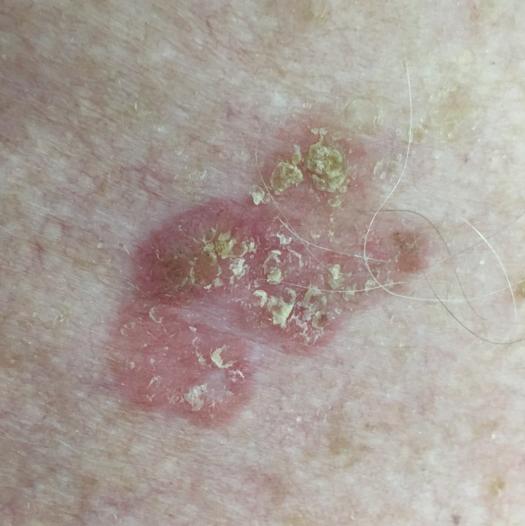}
        }
        \hspace*{0.05cm}
        \subfloat[MEL.]{
            \includegraphics[height=3.25cm, width=3.25cm]{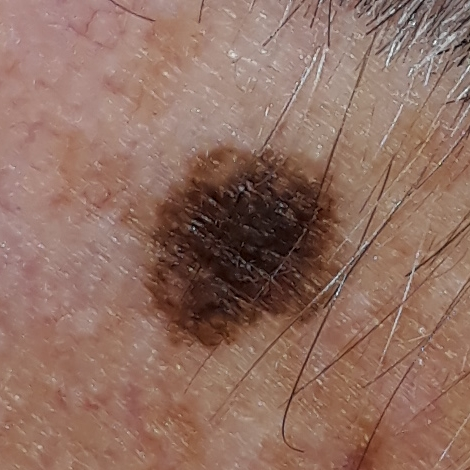}
        }
        \caption{Examples of images present in the PAD-UFES-20 dataset.}
        \label{fig:pad-samples}
    \end{figure}

    The dataset consists of 2,298 samples, distributed between 1,209 benign lesions (52.6\%) and 1,089 malignant lesions (47.4\%). In total, the dataset includes six types of skin lesions, and each sample consists of a clinical image acquired using a mobile device together with the patient's clinical information. The class distribution comprises three benign categories, corresponding to 730 samples of actinic keratosis (ACK), 235 of seborrheic keratosis (SEK), and 244 of melanocytic nevus (NEV), and three malignant categories, corresponding to 845 samples of basal cell carcinoma (BCC), 192 of squamous cell carcinoma (SCC), and 52 of melanoma (MEL). Figure~\ref{fig:pad-samples} shows a representative image sample for each type of lesion present in the dataset.



    The dataset includes $21$ clinical and sociodemographic characteristics of the patients. Among these variables are age, gender, anatomical location of the skin lesion, skin type according to the Fitzpatrick classification, and lesion diameter, represented by two features. In addition, qualitative information associated with the lesion is provided, including itching, growth, pain, change, bleeding, and elevation. The dataset also includes information on family history, considering father and mother, personal history of cancer, and history of skin cancer, as well as risk factors, represented by three features. Additionally, socioenvironmental information is provided, including the presence of piped water supply and a sewage system in the patient's residence. With the exception of age and lesion diameter, all complementary variables are categorical and, for this reason, were encoded using the one-hot encoding technique. At the end of this process, the complementary data were represented by a total of $81$ encoded features.

\subsection{NDB-UFES Dataset}

\begin{figure}[!htbp]
        \centering
        \subfloat[Squamous Carcinoma.]{
            \includegraphics[height=3.25cm, width=3.25cm]{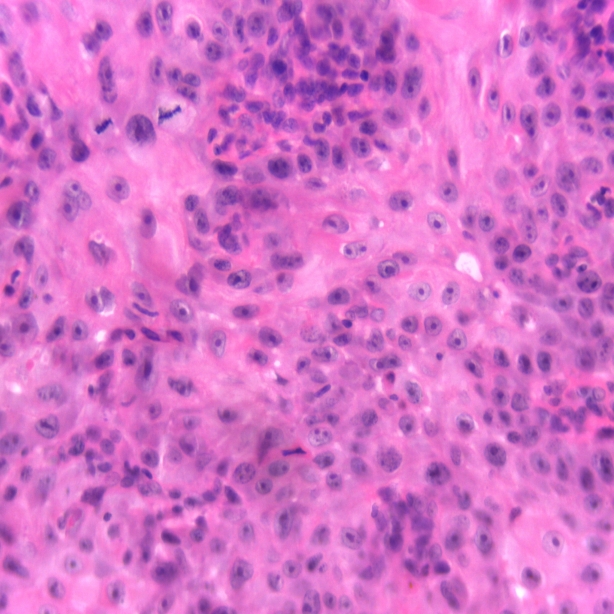}
        }
        \hspace*{0.05cm}
        \subfloat[Leukoplakia with dysplasia.]{
            \includegraphics[height=3.25cm, width=3.25cm]{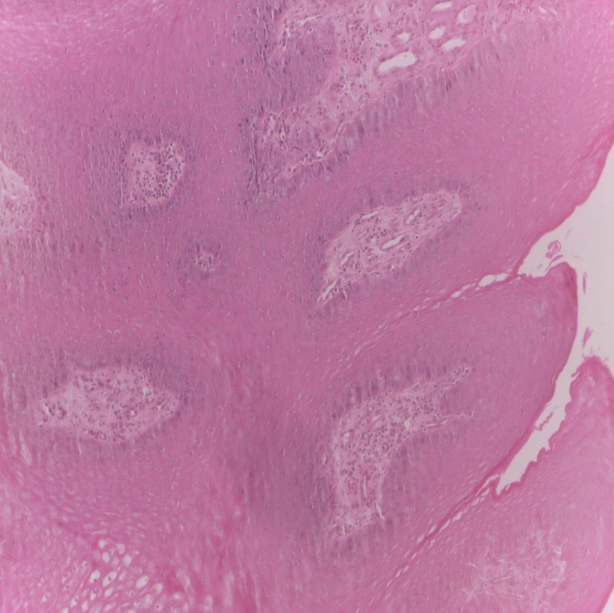}
        }
        \hspace*{0.05cm}
        \subfloat[Leukoplakia without dysplasia.]{
            \includegraphics[height=3.25cm, width=3.25cm]{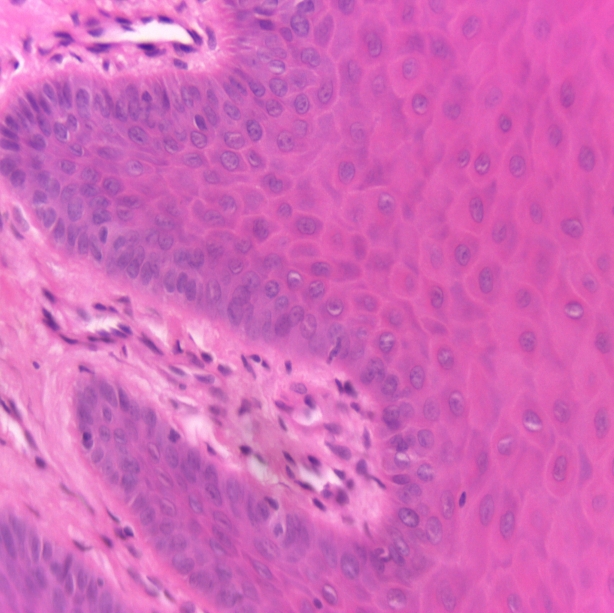}
        }
        \caption{Examples of images present in the NDB-UFES dataset.}
        \label{fig:ndb-samples}
    \end{figure}

    The second dataset analyzed in this study is NDB-UFES~\cite{de2023ndb}. In contrast to PAD-UFES-20, this dataset is composed of histopathological images of oral lesions as well as clinical and sociodemographic information from patients diagnosed with oral squamous cell carcinoma (OSCC), oral leukoplakia with dysplasia (LW/D), and oral leukoplakia without dysplasia (LW/oD). The dataset contains $237$ histopathological images, distributed across $91$ OSCC samples $(38.4\%)$, $89$ LW/D samples $(37.6\%)$, and $57$ LW/oD samples $(24.1\%)$. Figure~\ref{fig:ndb-samples} shows representative image samples for each type of oral lesion included in the dataset.


    The dataset contains $7$ clinical and sociodemographic characteristics of the patients. Among these variables are gender, age group, lesion location, and lesion size. In addition, sociodemographic and behavioral information is considered, including alcohol consumption habits, smoking habits, and regular sun exposure. With the exception of lesion size, all complementary variables are categorical and, for this reason, were encoded using the one-hot encoding technique. At the end of this process, the complementary data were represented by a total of $23$ encoded features.



\subsection{Implementation Details}
\label{subsec:impltedts}

    For evaluation, a holdout validation strategy was employed, in which $5/6$ of the dataset were allocated to training and the remaining $1/6$ was reserved for testing. In addition, 5-fold cross-validation was applied to the training subset, resulting in the training of $5$ distinct models. The reported results correspond to the mean and standard deviation of the metrics obtained by these $5$ models when evaluated on the test set. For the internal models, the default hyperparameters provided by the respective libraries were used, since ImageNet pre-trained weights were adopted. Specifically, for MambaVision, the pre-trained model MambaVision-L3-256-21K was adopted, whereas for VMamba, the VMamba-T model was used.


    For the external models, due to their greater configuration flexibility, a hyperparameter search was carried out using Optuna \cite{akiba2019optuna}, with the aim of maximizing the mean balanced accuracy (BACC). In this procedure, the $4/5$ folds obtained after applying cross-validation to the training set were used for model fitting, while the remaining $1/5$ fold was used for hyperparameter validation. This process was repeated until each fold had been used once as the validation set. For each external model, $50$ optimization trials were performed for each combination of internal model and dataset, with the objective of identifying the hyperparameter configuration that maximized BACC. The hyperparameter search space is listed in Table~\ref{tab:sens}.

    \begin{table}[ht]
        \renewcommand{\arraystretch}{1.3}
        \caption{Hyperparameter search space for Mambular and MambAttention, using VMamba or MambaVision as the internal classifier for both datasets.}
        \label{tab:sens}
        \centering
        \begin{tabular}{l|c|c}
        \hline 
        \textbf{Hyperparameter}                   & \textbf{Mambular}   & \textbf{MambAttention} \\ \hline
        learning rate ($lr$)                      & [0.00001 $\sim$ 0.01] & [0.00001 $\sim$ 0.01] \\
        dims model ($d_{model}$)                  & [32 $\sim$ 256] & [32 $\sim$ 256] \\
        number of layers ($n_{layers}$)           & [2 $\sim$ 10] & [2 $\sim$ 10] \\
        dims of conv  ($d_{conv}$)                & [2 $\sim$ 16] & [2 $\sim$ 16] \\
        attention layers ($n_{attention\ layer}$) & - & [1 $\sim$ 4] \\ \hline
        \end{tabular}
    \end{table}    

    The adopted loss function was weighted cross-entropy. Training was carried out over $150$ epochs, with a batch size of $30$, and an early stopping criterion was applied so that the process was interrupted in the absence of improvement for $15$ consecutive epochs. The optimizer employed was Adaptive Moment Estimation (Adam), with the initial learning rate found by Optuna and a weight decay of $0.001$. In addition, a reduce-on-plateau learning rate strategy was used, with a patience of $10$ epochs, a reduction factor of $0.1$, and a lower bound of $10^{-6}$ for the learning rate. The results were analyzed based on the mean and standard deviation of the metrics obtained for each evaluated combination. For performance evaluation, four metrics widely used in the literature were adopted: balanced accuracy (BACC), the weighted average of precision and recall, and area under the ROC curve (AUC).

    

\subsection{Results}
\label{subsec:results}

    After hyperparameter optimization using Optuna, each model was trained, validated, and subsequently evaluated on the test set. The experimental results for the PAD-UFES-20 dataset are listed in Table~\ref{table:mixed_fusion_pad_mamba}. For the NDB-UFES dataset, the corresponding results are listed in Table~\ref{table:mixed_fusion_ndb_mamba}. These tables summarize the performance obtained with the Mixed Fusion architecture, considering the combination of Mamba classifiers for visual and tabular processing, and compare it with the baseline \cite{Lima2026}, which uses approaches such as Transformers and CNNs.


    For the PAD-UFES-20 dataset, the best mean BACC did not surpass the baseline, reaching $0.7685 \pm 0.0291$ with the combination of VMamba and Mambular, compared with $0.7716 \pm 0.0267$ for the reference model. However, when considering the other metrics, namely precision, recall, and AUC, relevant gains were observed in relation to the baseline. In particular, for the recall metric, all combinations of the visual models and external classifiers outperformed the baseline, with the best result once again obtained by the combination of VMamba and Mambular, with a value of $0.8110 \pm 0.0166$, in contrast to the baseline value of $0.7600 \pm 0.0255$.


    \begin{table}[htp]
        \caption{Summary of evaluation of mixed fusion on PAD-UFES-20 dataset using a Mamba Classifier for clinical image. Mean and standard deviation of all evaluated metrics. Bold values have the best mean in table}
        \renewcommand{\arraystretch}{1.1}
        \begin{adjustbox}{width=\columnwidth,center}
            \centering
            \label{table:mixed_fusion_pad_mamba}
            \begin{tabular}{c|c|c|c|c|c}
                \hline
                \textbf{Internal classifier} & \textbf{External classifier} & \textbf{BACC} & \textbf{Precision} & \textbf{Recall} & \textbf{AUC} \\ 
                \hline
                \multirow{2}{*}{VMamba} 
                & Mambular      &  $0,7685 \pm 0,0291$ & $\mathbf{0,8193 \pm 0,0177}$ & $\mathbf{0,8110 \pm 0,0166}$ & $\mathbf{0,9548 \pm 0,0034}$ \\
                & MambAttention &  $0,7050 \pm 0,0115$ & $0,7748 \pm 0,0221$ & $0,8068 \pm 0,0108$ & $0,9358 \pm 0,0047$ \\
                \hline
                \multirow{2}{*}{MambaVision} %
                & Mambular      &  $0,7661 \pm 0,0147$ & $0,7834 \pm 0,0146$ & $0,7948 \pm 0,0071$ & $0,9472 \pm 0,0048$ \\
                & MambAttention &  $0,6983 \pm 0,0280$ & $0,7508 \pm 0,0155$ & $0,7791 \pm 0,0075$ & $0,9385 \pm 0,0022$ \\
                \hline
                \multicolumn{6}{c}{\textbf{Baseline \cite{Lima2026}}} \\ \hline
                ViT  & NN       & $\mathbf{0.7716 \pm 0.0267}$ & $0.8080 \pm 0.0228$ & $0.7600 \pm 0.0255$ & $0.9394 \pm 0.0047$ \\
                \hline
            \end{tabular}
        \end{adjustbox}
    \end{table}

    For the NDB-UFES dataset, the combination of MambaVision and MambAttention outperformed the baseline across all evaluated metrics. The best mean BACC was $0.8281 \pm 0.0367$, compared with $0.7780 \pm 0.0626$ obtained by the baseline. The recall metric also showed improvement over the baseline in all tested models, reaching $0.8667 \pm 0.0299$, whereas the baseline achieved $0.7940 \pm 0.0428$. For the precision metric, a mean of $0.8850 \pm 0.0222$ was observed, and for the AUC metric, a mean of $0.9790 \pm 0.0116$ was obtained.

    \begin{table*}[ht]
        \caption{Summary of evaluation of mixed fusion on NDB-UFES dataset using a Mamba Classifier for the histopathological image. Mean and standard deviation of all evaluated metrics. Bold values have the best mean in table}
        \renewcommand{\arraystretch}{1.1}
        \begin{adjustbox}{width=\columnwidth,center}
            \centering
            \label{table:mixed_fusion_ndb_mamba}
            \begin{tabular}{c|c|c|c|c|c}
                \hline
                \textbf{Internal classifier} & \textbf{External classifier} & \textbf{BACC} & \textbf{Precision} & \textbf{Recall} & \textbf{AUC} \\ 
                \hline
                \multirow{2}{*}{VMamba} 
                & Mambular      & $0,8059 \pm 0,0722$ & $0,8394 \pm 0,0659$ & $0,8308 \pm 0,0699$ & $0,9493 \pm 0,0174$\\
                & MambAttention & $0,8119 \pm 0,0089$ & $0,8616 \pm 0,0155$ & $0,8513 \pm 0,0103$ & $0,9636 \pm 0,0118$ \\
                \hline
                \multirow{2}{*}{MambaVision} 
                & Mambular      & $0,7941 \pm 0,0178$ & $0,8444 \pm 0,0260$ & $0,8308 \pm 0,0205$ & $0,9476 \pm 0,0167$\\
                & MambAttention & $\mathbf{0,8281 \pm 0,0367}$ & $\mathbf{0,8850 \pm 0,0222}$ & $\mathbf{0,8667 \pm 0,0299}$ & $\mathbf{0,9790 \pm 0,0116}$\\
                \hline
                \multicolumn{6}{c}{\textbf{Baseline \cite{Lima2026}}} \\ \hline
                CoaT & LightGBM & $0.7780 \pm 0.0626$ & $0.8080 \pm 0.0455$ & $0.7940 \pm 0.0428$ & $0.9104 \pm 0.0335$ \\
                \hline
            \end{tabular}
        \end{adjustbox}
    \end{table*}

Mamba-based models, together with the mixed fusion approach, showed superior performance in comparison with the approach based on Transformers and CNNs, although BACC was lower than the baseline on the PAD-UFES-20 dataset. One of the main advantages of the Mixed Fusion architecture lies in the possibility of performing explainability analyses through the Shapley Additive Explanations (SHAP) method \cite{scott2017unified}, which contributes to a more detailed understanding of the model’s decision-making process.


The configuration of the SHAP experiments was defined as follows: the model corresponding to the best fold from the training set was used, selected based on the performance observed on the validation set, and applied TreeSHAP to compute the values. The presented results correspond to the SHAP values calculated for all samples in the test set.

\begin{figure}[htp]
    \centering
    \subfloat[Samples of class SCC on PAD-UFES-20]{\includegraphics[scale=0.29]{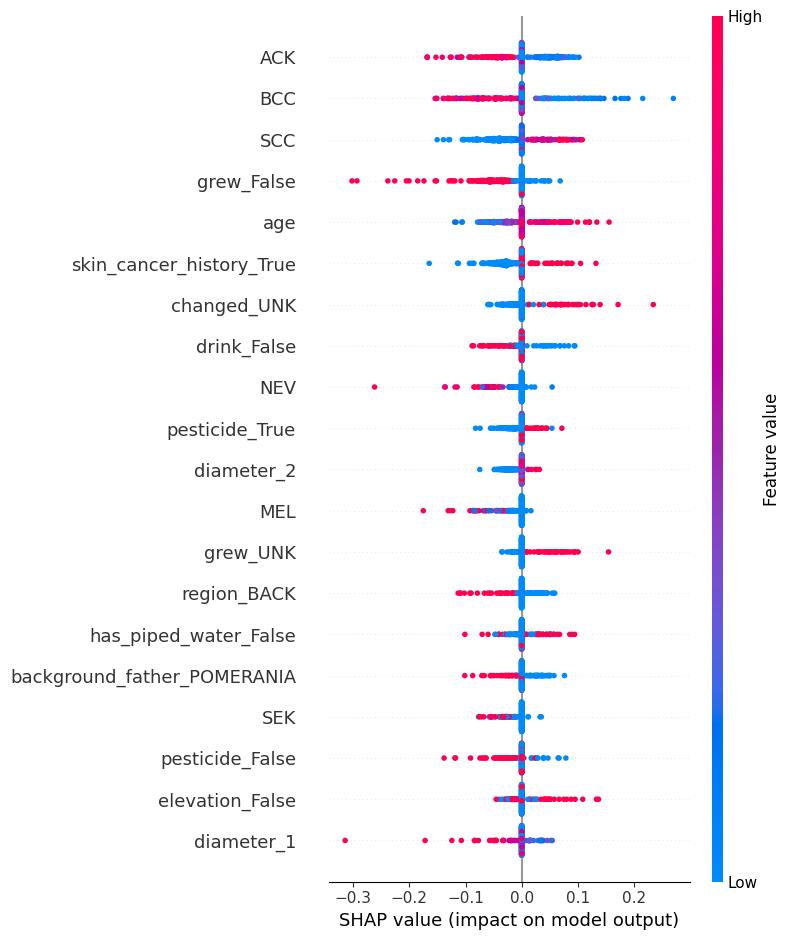}
    \label{fig:shap_pad_SCC}
    } 
    \hfill
    \subfloat[Samples of class OSCC on NDB-UFES]{\includegraphics[scale=0.29]{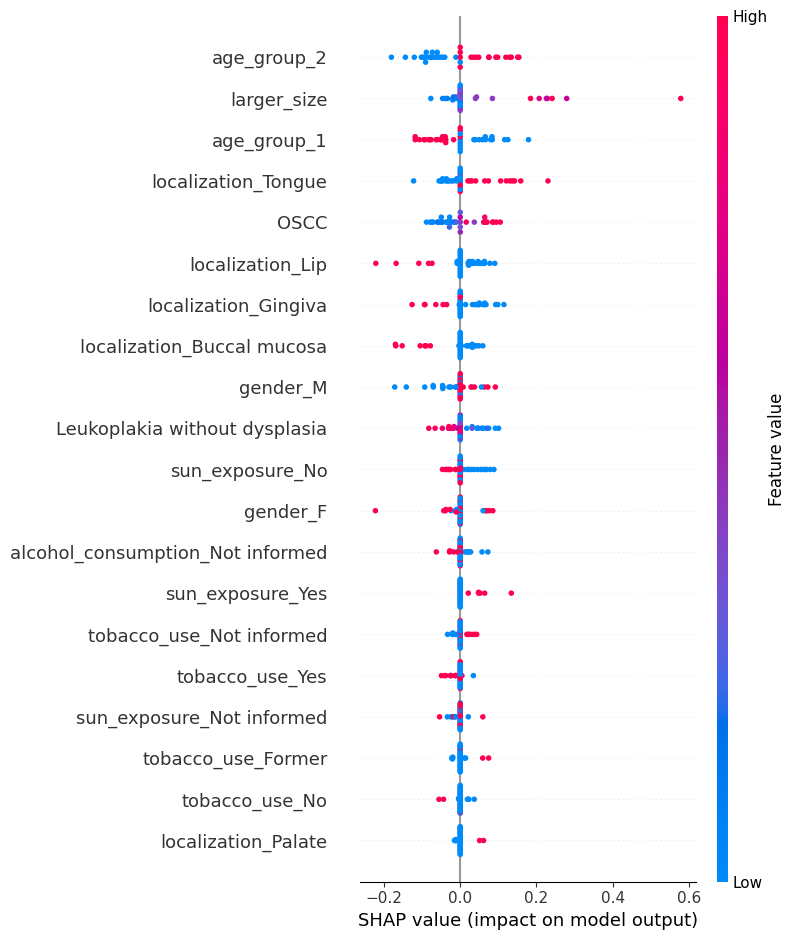}
    \label{fig:shap_NDB_OSCC}
    }
    \caption{Feature impact on model output for classes on PAD-UFES-20 using VMamba/Mambular and NDB-UFES using MambaVision/MambAttention.}
    \label{fig:shap_pad_no_poincare_skincancer}
\end{figure}

For the PAD-UFES-20 dataset, the SHAP analysis was performed considering the combination of the VMamba internal classifier and the Mambular external classifier. Figure~\ref{fig:shap_pad_SCC} deploys, for the true SCC class, the impact of the 20 most relevant features on the model output, ordered according to the mean absolute value of the SHAP values. The features referring to the dataset classes (``BCC'', ``SCC'', ``MEL'', ``ACK'', ``NEV'' and ``SEK'') correspond to the probabilities estimated by VMamba for each class, obtained solely from the information contained in the image.

It can be observed that the information associated with image classification that appears most frequently among the most relevant features generally corresponds to the classes that show the greatest confusion with SCC, in agreement with Pacheco et al.~\cite{pacheco2020impact}. In addition to these variables, features related to lesion growth (``grew''), changes in its shape (``changed''), and patient age (``age'') also stand out, being aligned with clinical criteria used in the 7-point checklist \cite{johr2002dermoscopy, Congdon2023}. Additionally, lesion diameter (``diameter\_1'' and ``diameter\_2''), which is also related to the ABCDE rule widely used by dermatologists in the assessment of skin lesions, was identified as relevant \cite{duarte2021clinical, wu2022skin}. It is also worth highlighting the presence of skin cancer history (``skin\_cancer\_history'') and patient ancestry (``background\_father\_POMERANIA''), which may be related to genetic, environmental, or socioeconomic factors.


For the NDB-UFES dataset, the SHAP analysis was performed considering the combination of the MambaVision internal classifier and the MambAttention external classifier. Figure \ref{fig:shap_NDB_OSCC} deploys, for the true OSCC class, the impact of the 20 most relevant features on the model output, ordered according to the mean absolute value of the SHAP values. The features referring to the dataset classes (``Leukoplakia without dysplasia'', ``Leukoplakia with dysplasia'', and ``OSCC'') correspond to the probabilities estimated by MambaVision for each class, obtained solely from the information contained in the image.


The results of the SHAP analysis indicate that age group exerted a relevant influence on the prediction of OSCC, with particular emphasis on the groups between 41 and 60 years old (``age\_group\_1'') and above 60 years old (``age\_group\_2''), both appearing among the most important features. This finding is consistent with the literature, which reports a higher prevalence of OSCC, especially in the tongue region, in men over 50 years of age \cite{del2019oral, andrade2015associated, tan2023oral}. The presence of other relevant variables can also be observed, such as lesion location (for example, ``localization\_Lip''), lesion size (``larger\_size''), gender, alcohol consumption (``alcohol\_consumption''), and tobacco use (``tabacco\_use''), factors widely recognized in clinical practice as being associated with the diagnosis of this neoplasm.


When directly comparing the results obtained in this study with those reported by Lima et al.~\cite{Lima2026}, differences can be observed in the explainability pattern of the models. In the present work, the Mamba-based models distributed predictive importance across a broader set of variables, including sociodemographic and clinical factors, which suggests a more multifactorial explanatory pattern. In contrast, in the analysis reported for the baseline, the explanation was more concentrated on a restricted set of variables, resulting in greater magnitude of the SHAP values associated with these attributes. For the SCC class in the PAD-UFES-20 dataset, the baseline mainly highlighted lesion diameter, age, and variables related to the clinical classes. For the OSCC class in the NDB-UFES dataset, a predominance of the Leukoplakia with dysplasia class was observed, followed by features associated with anatomical location and smoking. Thus, whereas the baseline appears to adopt a more selective predictive strategy, strongly anchored in a reduced number of markers, the Mamba-based approach operates from a more distributed composition of evidence.



\section{Conclusion}
\label{sec:conclusion}

    We present a complementary approach for the integration of multimodal medical data, based on the joint use of a state space model, represented by the Mamba architecture, applied to visual and tabular data through the Mixed Fusion architecture. The experimental results indicate that the use of Mamba provided consistent gains in the recall metric across both evaluated datasets when compared with Transformer-based approaches. However, in the PAD-UFES-20 dataset, the approach showed slightly lower performance in terms of BACC, whereas superior results were obtained in the NDB-UFES dataset. The application of the SHAP method, with the aim of analyzing the model’s decision-making process, indicate that the most relevant variables are aligned with criteria commonly considered by specialists in clinical practice. In addition, it was observed that the model explanation is distributed across a broader set of variables, including sociodemographic and clinical factors, which suggests a more multifactorial explanatory pattern. As future work, we intend to expand this investigation through the integration of other explainability techniques, enabling the comparison of different approaches, the identification of new interpretative patterns, and the potential increase of the model’s applicability in clinical settings.



\begin{credits}
\subsubsection{\ackname} This study was financed in part by the FAPES, Brazil - Process No. 2023-CSN92, through a PhD scholarship awarded to M.B. Rocha; R.A. Krohling thanks the Brazilian research agency CNPq, Brazil – grant no. 302021/2025-6; 
The authors also thank the PROAP/CAPES (Portaria no. 206, 9/4/2018).
The funder had no role in study design, data collection and analysis, decision to publish, or preparation of the manuscript.

\subsubsection{Data and code availability} PAD-UFES-20~\cite{pacheco2020pad} and NDB-UFES~\cite{de2023ndb} are publicly available datasets. The source code is public available at GitHub at \url{https://github.com/MatheusBecali/mamba-mixed-fusion}.

\subsubsection{\discintname}
The authors have no competing interests to declare that are relevant to the content of this article.
\end{credits}
%


%
%
\bibliographystyle{splncs04}
\bibliography{bibliography}
%




\end{document}